\documentclass{article}

\usepackage{amsmath,amsfonts,bm}

\def\eqref#1{equation~\ref{#1}}

\def\1{\bm{1}}

\def\vv{{\bm{v}}}

\def\vx{{\bm{x}}}

\DeclareMathAlphabet{\mathsfit}{\encodingdefault}{\sfdefault}{m}{sl}
\SetMathAlphabet{\mathsfit}{bold}{\encodingdefault}{\sfdefault}{bx}{n}

\DeclareMathOperator*{\argmin}{arg\,min}

\usepackage{microtype}
\usepackage{graphicx}
\usepackage{subfigure}
\usepackage{booktabs}
\usepackage{epsfig}
\usepackage{graphicx}
\usepackage{amsmath}
\usepackage{amssymb}
\usepackage{tabu}
\usepackage{tabularx}
\usepackage{multirow}
\usepackage{subfigure}
\usepackage{booktabs} 

\usepackage{hyperref}

\usepackage[accepted]{icml2020}

\icmltitlerunning{Deflecting Adversarial Attacks}

\begin{document}

\twocolumn[
\icmltitle{Deflecting Adversarial Attacks}




\begin{icmlauthorlist}
\icmlauthor{Yao Qin}{uni,goo}
\icmlauthor{Nicholas Frosst}{goo}
\icmlauthor{Colin Raffel}{goo}
\icmlauthor{Garrison Cottrell}{uni}
\icmlauthor{Geoffrey Hinton}{goo}
\end{icmlauthorlist}

\icmlaffiliation{uni}{University of California, San Diego}
\icmlaffiliation{goo}{Google Brain}

\icmlcorrespondingauthor{Yao Qin}{yaoqin@google.com}
\icmlcorrespondingauthor{Colin Raffel}{craffel@google.com}
\icmlcorrespondingauthor{Garrison Cottrell}{gary@ucsd.edu}
\icmlcorrespondingauthor{Geoffrey Hinton}{geoffhinton@google.com}

\icmlkeywords{Machine Learning, ICML}

\vskip 0.3in
]



\printAffiliationsAndNotice{}  

\begin{abstract}
There has been an ongoing cycle where stronger defenses against adversarial attacks are subsequently broken by a more advanced defense-aware attack. We present a new approach towards ending this cycle where we ``deflect'' adversarial attacks by causing the attacker to produce an input that semantically resembles the attack's target class. To this end, we first propose a stronger defense based on Capsule Networks that combines three detection mechanisms to achieve state-of-the-art detection performance on both standard and defense-aware attacks. We then show that undetected attacks against our defense often perceptually resemble the adversarial target class by performing a human study where participants are asked to label images produced by the attack. These attack images can no longer be called ``adversarial'' because our network classifies them the same way as humans do.
\end{abstract}

\section{Introduction}

\begin{figure}[t]
  \centering
  \includegraphics[width=\linewidth]{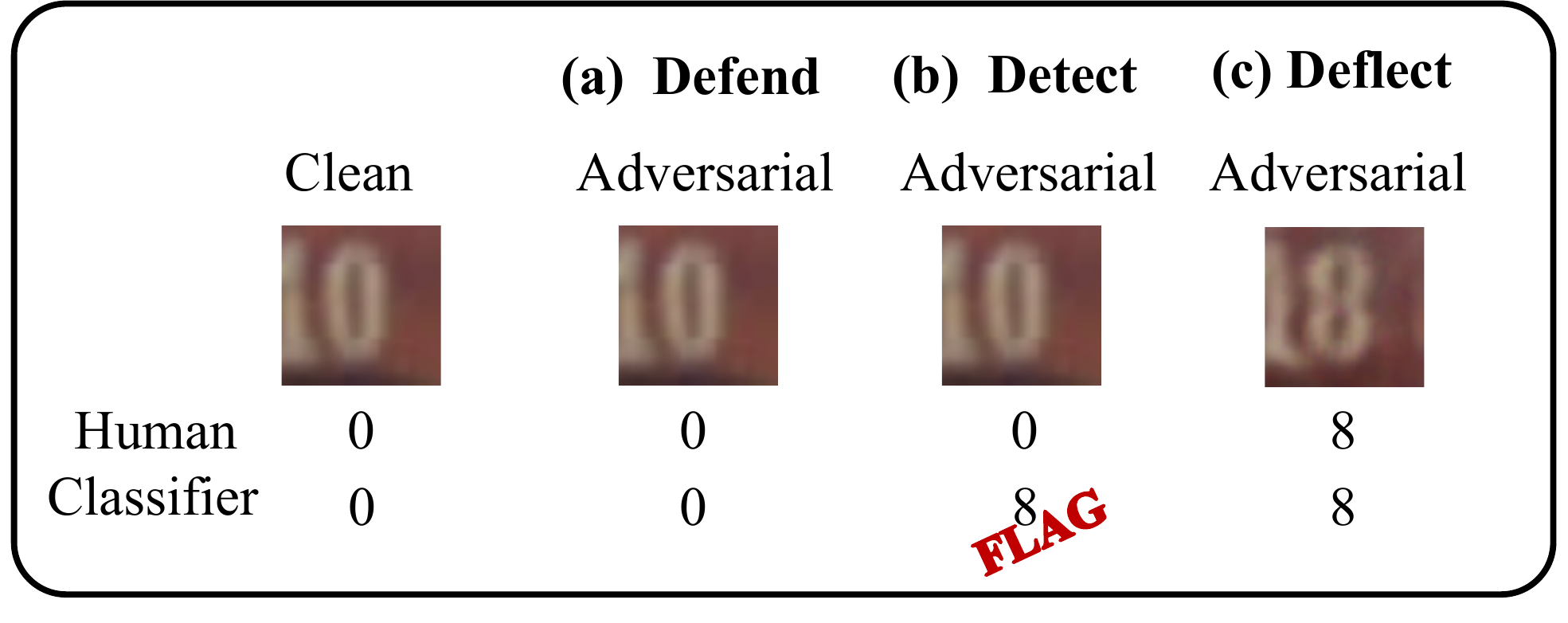}
  \vspace{-7mm}
  \caption{Different results of an adversarial attack against three different defense approaches. The original class is 0 and the adversarial target class is 8.}\label{fig:defense}
  \vspace{-5mm}
\end{figure}

\begin{figure}[t]
  \centering
  \includegraphics[width=\linewidth]{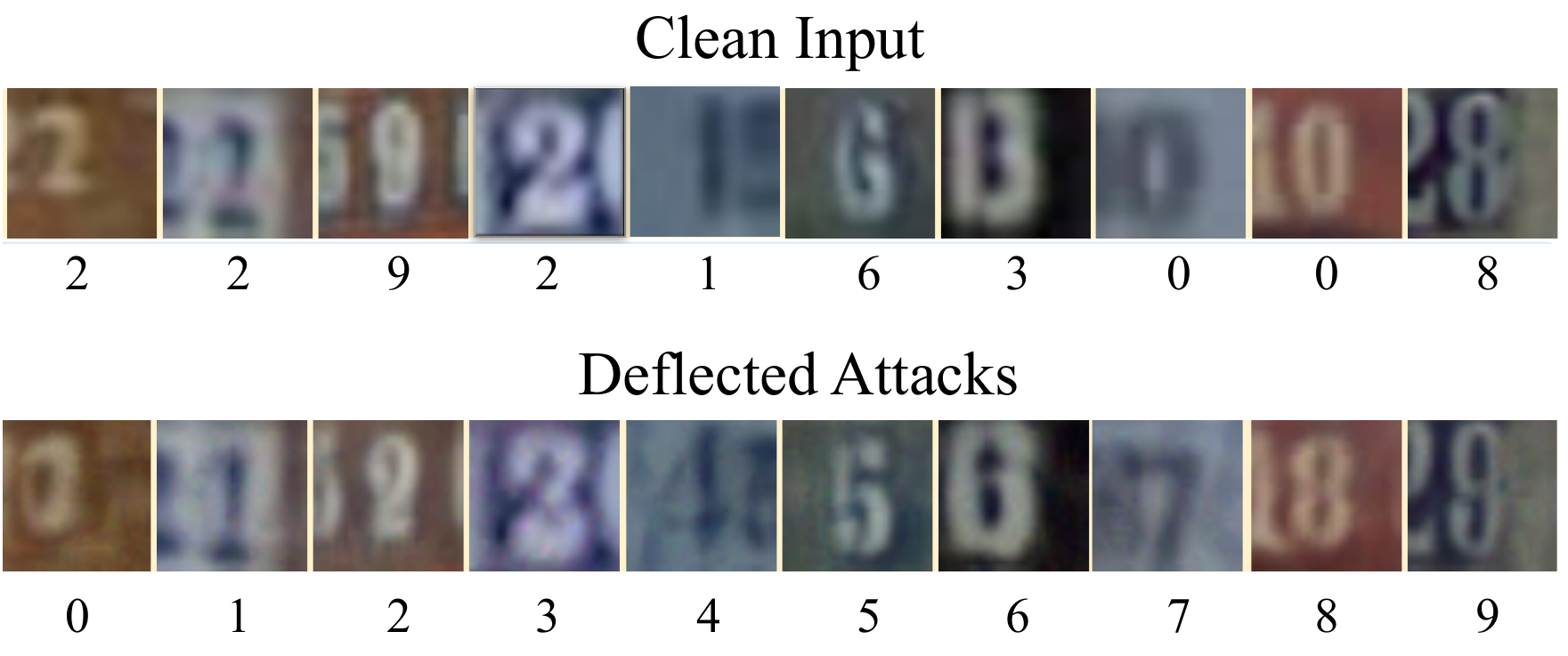}
  \vspace{-7mm}
  \caption{Deflected adversarial attacks on the SVHN dataset. These images were generated by a defense aware attack and the maximal adversarial perturbation is bounded by 16/255.}\label{fig:svhn_deflect}
\end{figure}

Adversarial attacks have been the subject of constant research since they were first discovered~\cite{szegedy2013, goodfellow2014,kurakin2016}. Most of this research has been focused on the creation of more robust models to \textbf{defend} against adversarial attacks~\cite{song2017pixeldefend, madry2017towards, yang2019me,goodfellow2018evaluation}, where the input image is correctly classified as the original class rather than the adversarial target class, as illustrated in Figure~\ref{fig:defense} (a). However, better defenses have led to the development of stronger attack algorithms to break these defenses~\cite{madry2017towards, carlini2017towards, chen2018ead, athalye2018obfuscated}. After several iterations of creating and breaking defenses, some research focused on adversarial attack \textbf{detection}~\cite{grosse2017statistical, feinman2017detecting, metzen2017detecting, lee2018simple, qin2019detecting, roth2019odds}. Detection algorithms aim to distinguish adversarial attacks from real data and then flag the adversarial input, instead of attempting to correctly classify such inputs, as shown in Figure~\ref{fig:defense} (b). However, this strategy fell into the same creating/breaking cycle: Many state-of-the-art methods~\cite{roth2019odds, ma2018characterizing, lee2018simple} claiming to detect adversarial attacks were broken shortly after publication with a defense-aware attack~\cite{hosseini2019odds, carlini2017adversarial, athalye2018obfuscated}. We attempt to get ahead of this cycle by focusing on the \textbf{deflection} of adversarial attacks, shown in Figure~\ref{fig:defense} (c): If the result of the adversarial optimization of an image looks to a human like the adversarial target class rather than its original class, then the image can hardly be called adversarial anymore. We call such attacks ``deflected''. Some examples are shown in Figure~\ref{fig:svhn_deflect}.

In this paper, we propose a network and detection mechanism based on Capsule layers~\cite{sabour2017,qin2019detecting} that either detects attacks accurately or, for undetected attacks, often pressures the attacker to produce images that resemble the target class (thereby deflecting them).
Our network architecture is made up of two components: A capsule classification network that classifies the input, and a reconstruction network that reconstructs the input conditioned on the pose parameters of the predicted capsule. Apart from the classification loss and $\ell_2$ reconstruction loss used in~\cite{sabour2017, qin2019detecting}, we introduce an extra cycle-consistency training loss which constrains the classification of the winning capsule reconstruction to be the same as the classification of the original input. This new auxiliary training loss encourages the reconstructions to more closely match the class-conditional distribution and helps the model detect and deflect adversarial attacks. 

In addition, we propose two new attack-agnostic detection methods based on the discrepancy between the winning-capsule reconstruction of clean and adversarial inputs. We find that a detection method that combines ours with the one proposed by \cite{qin2019detecting} performs best. We show that this method can accurately detect white-box and black-box attacks based on three different distortion metrics (EAD~\cite{chen2018ead}, CW~\cite{carlini2017towards} and PGD~\cite{madry2017towards}) on both the SVHN and CIFAR-10 datasets. Following the suggestions in~\cite{athalye2018obfuscated, carlini2017adversarial}, we also propose defense-aware attacks for our new detection method. We find that our detection methods significantly outperform state-of-the-art methods on defense-aware attacks. Finally, we perform a human study to verify that many of the undetected adversarial attacks against our model have been successfully deflected, i.e.\ adversarial images from both defense-aware and standard attacks against our detection mechanism are frequently classified as the target class by humans. In contrast, successful attacks against baseline models do not have this property.
To summarize, our main contributions are as follows:
\begin{itemize}
\item We introduce the notion of \textbf{deflecting adversarial attacks}, which presents a step towards ending the battle between attacks and defenses.
\item We propose a new cycle-consistency loss which trains a CapsNet to encourage the winning-capsule reconstruction to closely match the class-conditional distribution and show that this can help detect and deflect adversarial attacks.
\item We introduce two attack-agnostic detection methods based on the discrepancy between the winning-capsule reconstruction of the clean and adversarial inputs, and design a defense-aware attack to specifically attack our detection mechanisms.
\item We show through extensive experiments on SVHN and CIFAR-10 that our detection mechanism can achieve state-of-the-art performance in detecting white-/black-box standard and defense-aware attacks.
\item We perform a human study to show that our approach, unlike previous methods, is able to deflect a large percentage of undetected adversarial attacks.
\end{itemize}

\begin{figure*}[t]
  \centering
  \includegraphics[width=0.7\linewidth]{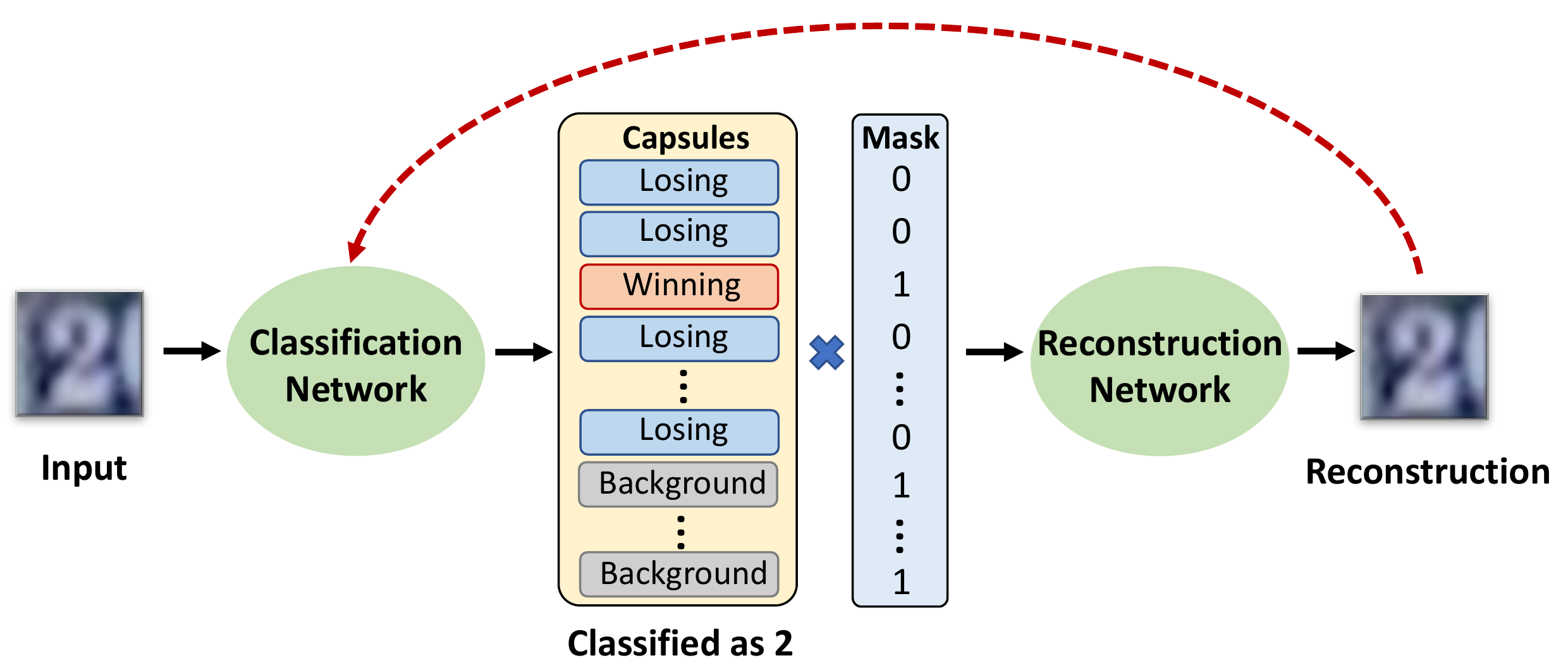}
  \vspace{-3mm}
  \caption{The network architecture with cycle-consistent winning capsule reconstructions.}\label{fig:arch1}  
  \vspace{-3mm}
\end{figure*}
\vspace{-4mm}
\section{Network Architecture}\label{sec:arch}
In order to design a model that is strong enough to deflect adversarial attacks, we build our network based on CapsNet~\cite{sabour2017}. Figure~\ref{fig:arch1} shows the pipeline of our network architecture. The final layer of our classifier is a Capsule layer (``CapsLayer'' for short) which includes both class capsules and background capsules. These capsules are intended to encode feature attributes corresponding to the class and the background respectively. Given an input $\vx$, the output of a CapsLayer is a prediction $f(\vx)$ and a pose parameter $\vv$ for all the classes and the background, where $\vv_i$ denotes the pose parameter for class $i$. As in the initial Capsules proposed in~\cite{sabour2017}, the magnitude of the activation vector of a capsule encodes the existence of an instance of the class and the orientation of the activation vector encodes instantiation parameters of the instance, such as its pose.  Therefore, the magnitudes of the capsules' activations are used to perform classification while the activation vector of the winning class capsule together with the activation vectors of the background capsules are used as the input to the reconstruction network. We use $r(\vv_{i=f(\vx)})$ and $r(\vv_{i\neq f(\vx)})$ to represent the  reconstruction from the winning capsule and a losing capsule respectively. The reconstruction network uses the activations of all the background capsules as well as the activation of one class capsule but we omit this to simplify the notation. More details of the network architecture used in this paper are provided in Supplementary Material.
\vspace{-3mm}
\paragraph{Cycle-consistent winning-capsule reconstructions}
The CapsNet~\cite{sabour2017} is trained with two loss terms: a marginal loss for the classification and an $\ell_2$ reconstruction loss. To encourage the reconstruction to more closely match the class conditional distribution and help the model detect and deflect adversarial attacks, we additionally incorporate an extra cycle-consistency loss $\ell_{cyc}$ which constrains the reconstruction from the winning capsule to be classified as the same class as the input, formulated as:
\begin{equation}
\ell_{cyc} = \ell_{net}(f(r(\vv_{i=f(\vx)})), f(\vx)),
\end{equation}
where $\ell_{net}$ is the cross-entropy loss function and $i \in \{0, 1, \dots, n \}$, $n$ denotes the number of classes in the dataset. This can be achieved by feeding the reconstruction corresponding to the winning capsule back into the classification network, shown as the dotted red line in Figure~\ref{fig:arch1}. This extra training loss together with our Cycle-consistent Detector (introduced in Section~\ref{sec:ccd}) can help detect adversarial attacks. In addition, since the winning-capsule reconstructions are optimized to more closely match the class conditional data distribution, it becomes easier for our model to deflect adversarial attacks. 
\vspace{-3mm}
\section{Detection Methods}
\begin{figure*}[t]
  \centering
  \includegraphics[width=0.8\linewidth]{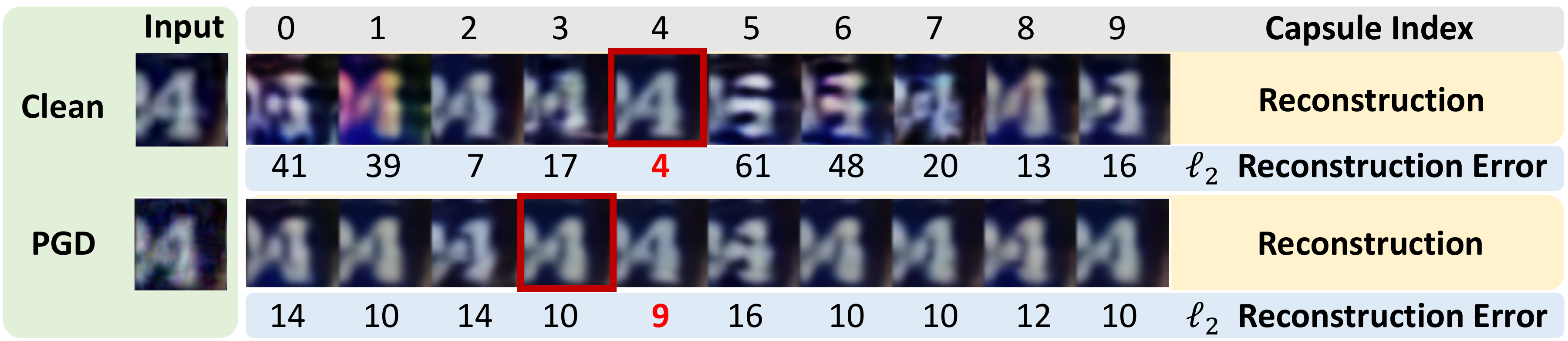}
  \vspace{-1mm}
  \caption{An example of a clean input, an adversarial example generated via a PGD attack, and the reconstructions for the clean and adversarial inputs from each class capsule.  The reconstruction corresponding to the winning capsule is surrounded by a red box. Under each reconstruction is its $\ell_2$ reconstruction error; the smallest reconstruction error is highlighted in {\color{red}red}. Both the clean input and its winning capsule reconstruction are classified as `4'. The PGD attack is classified as the target class `3' but its winning capsule reconstruction is classified as `4'.}\label{fig:exam1}
\end{figure*}

\vspace{-2mm}
In this paper, we use three reconstruction-based detection methods to detect standard attacks. They are: \textbf{G}lobal \textbf{T}hreshold \textbf{D}etector (GTD), first proposed in~\cite{qin2019detecting}, \textbf{L}ocal \textbf{B}est \textbf{D}etector (LBD) and \textbf{C}ycle-\textbf{C}onsistency \textbf{D}etector (CCD).
\vspace{-2mm}
\paragraph{Global Threshold Detector}
When the input is adversarially perturbed, the classification given to the input may be incorrect, but the reconstruction is often blurry and therefore the distance between the adversarial input and the reconstruction is larger than would be expected from normal input. This allows us to detect the input as adversarial with the Global Threshold Detector. This method, proposed in \cite{qin2019detecting}, measures the reconstruction error between the input and its reconstruction from the winning capsule. If the reconstruction error is greater than a global threshold $\theta$: 
\begin{equation}
\lVert r(\vv_{i=f(\vx)}) - \vx \rVert_2 > \theta,
\end{equation}
then the input is flagged as an adversarial example. 
\vspace{-2mm}
\paragraph{Local Best Detector}
When the input is a clean image, the reconstruction error from the winning capsule is smaller than that of the losing capsules, where an example is shown in the first row of Figure~\ref{fig:exam1}. This is likely because the $\ell_2$ reconstruction objective only minimizes the reconstruction from the winning capsule during training. However, when the input is an adversarial example, the reconstruction from the  capsule corresponding to the correct label can be even closer to the input compared to the reconstruction corresponding to the winning capsule (see the second row in Figure~\ref{fig:exam1}). Therefore, we propose the ``Local Best Detector'' (LBD) to detect such adversarial images whose reconstruction error from the winning capsule is not the smallest:
\begin{equation}
\argmin_j \lVert r(\vv_{j}) - \vx \rVert_2  \neq f(\vx), \quad j \in \{0, 1, \dots, n \},
\end{equation}
where $n$ is the number of classes in the dataset.
\paragraph{Cycle-Consistency Detector}\label{sec:ccd}
If the input is a clean image, the reconstruction from the winning capsule will resemble the input. Our model should ideally assign the same class to the reconstruction of the winning capsule as the clean input. This behavior is reinforced by training with the cycle-consistency loss. For example, as shown in Figure~\ref{fig:exam1} both the clean input and its winning-capsule reconstruction are classified as 4. However, when the input is an adversarial example that is perceptually indistinguishable from the clean image but pressures the model to predict the target class, the reconstruction of the winning capsule often appears closer to the clean input and/or is blurry. As a result, the reconstruction of the winning capsule is often not classified as the target class. As shown in Figure~\ref{fig:exam1}, the adversarial input has been classified as the target class ``3'' while the reconstruction corresponding to the winning capsule is classified as ``4''. Therefore, the Cycle-Consistency Detector (CCD) is designed to flag the input as an adversarial example if the input $\vx$ and its reconstruction of the winning capsule $r(\vv_{i=f(\vx)})$ are not classified as the same class:
\begin{equation}\label{eqn:cc}
f(r(\vv_{i=f(\vx)})) \neq f(\vx).
\end{equation}

In this paper, we use these three detectors together to detect adversarial examples. In other words, we flag any input as adversarial if it's classified as adversarial by any of the detection mechanisms. As a result, an adversarial input can only go undetected if it passes all three detection mechanisms. 
\vspace{-2mm}
\section{The Defense-Aware CC-PGD Attack}
\label{sec:ccpgd}

In order for an attack mechanism to generate an adversarial example $\vx' = \vx + \Delta$ (where $\Delta$ is a small adversarial perturbation) that can both cause a misclassification and is not detected by our detection mechanisms, the constructed adversarial attack must:

\begin{itemize}
\item successfully fool the classifier: $f(\vx') = t$ and $f(\vx)\neq t$, where $t$ is the target class.
\item avoid being detected by the Global Threshold Detector (GTD), the attack needs to constrain the reconstruction of the winning capsule to be close to the input.
\item fool the Local Best Detector (LBD), the attack should encourage the reconstructions from all the losing capsules to be far away from the input to ensure the reconstruction error of the winning capsule is the smallest.
\item circumvent the Cycle-Consistency Detector (CCD) by fooling the classifier into making the target prediction when it is fed the winning-capsule reconstruction of the adversarial input, that is: $f(r(\vv_{i=f(\vx')})) = f(\vx') = t$.
\end{itemize}

To generate such an attack, we follow~\cite{qin2019detecting} and devise attacks which consist of two stages at each gradient step. The first stage attempts to fool the classifier by following a standard attack (e.g., a standard PGD attack) which follows the gradient of the cross-entropy loss function with respect to the input. Then, in the second stage, we focus on fooling the detection mechanisms by taking the reconstruction error and cycle-consistency into consideration. This can be formulated as minimizing the reconstruction loss $\ell_{r}$, which consists of three components: the reconstruction loss corresponding to the Global Threshold Detector $\ell_{g}$, the reconstruction loss corresponding to the Local Best Detector $\ell_{l}$ and the cycle-consistency classification loss corresponding to the Cycle-Consistency Detector $\ell_{cyc}$. Specifically, the reconstruction loss is defined as:
\begin{equation}\label{loss}
\begin{split}
\ell_{r}(\vx') & = \alpha_1 \cdot  \ell_{g}(\vx') + \alpha_2 \cdot \ell_{l}(\vx') + \alpha_3 \cdot \ell_{cyc}(\vx')\\
& = \alpha_1 \cdot \lVert r(\vv_{i=f(\vx')}) - \vx' \rVert_2  \\
& \quad \quad - \alpha_2 \cdot  \frac{\sum_{k \neq f(\vx')}^{n}\lVert r(\vv_{k}) - \vx' \rVert_2}{n-1} \\ 
&  \quad \quad \quad \quad + \alpha_3 \cdot \ell_{net}(f(r(\vv_{i=f(\vx')})), f(\vx'))
\end{split}
\end{equation}
where $\vx' = \vx+\Delta$ is the adversarial example, $n$ is the number of the classes in the dataset, $\lVert r(\vv_{i=f(\vx')}) - \vx' \rVert_2$ is the winning-capsule reconstruction error and $\lVert r(\vv_{k\neq f(\vx')}) - \vx' \rVert_2$ is the losing-capsule reconstruction error. The hyperparameters $\alpha_1$, $\alpha_2$ and $\alpha_3$ are used to balance the importance of attacking each detector. Then, the adversarial perturbation can be updated in the second stage as:
\begin{equation}\label{stage2}
\Delta \leftarrow \textnormal{clip}_{\epsilon_\infty} (\Delta - c\cdot \textnormal{sign} (\nabla_\Delta (\ell_r(\vx+\Delta))),
\end{equation}
where $\epsilon_\infty$ is the $\ell_\infty$ norm bound and $c$ is the step size in each iteration. 
\begin{figure*}[t]
\minipage{0.248\textwidth}
  \includegraphics[width=\linewidth]{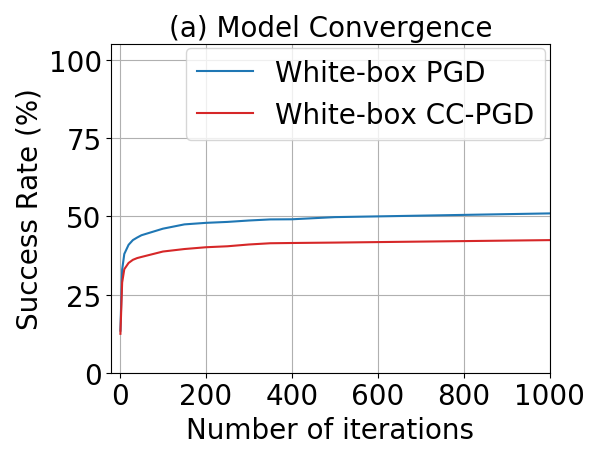}
\endminipage\hfill
\minipage{0.248\textwidth}
  \includegraphics[width=\linewidth]{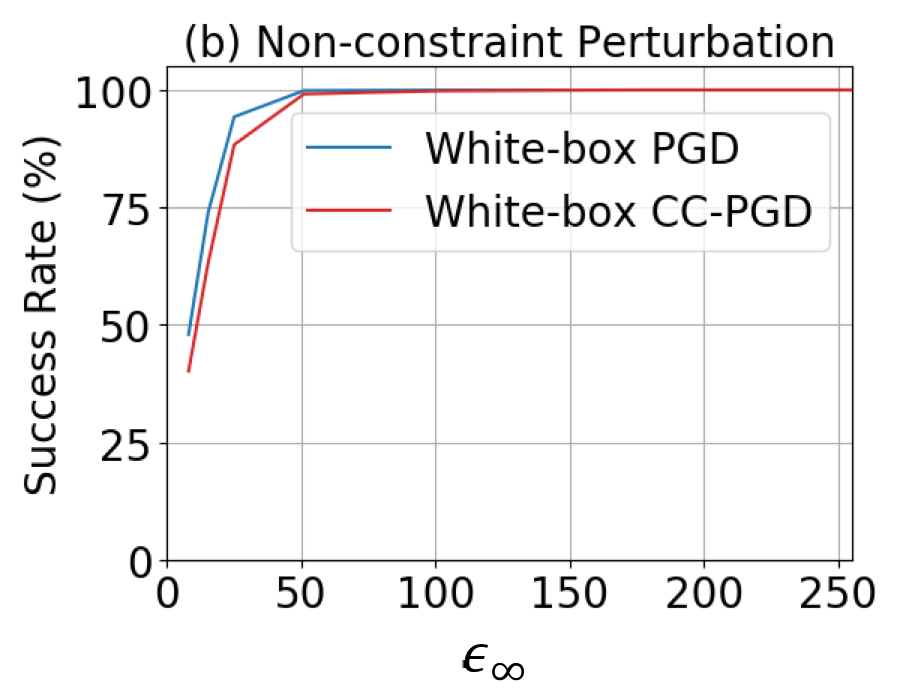}
\endminipage\hfill
\minipage{0.248\textwidth}
  \includegraphics[width=\linewidth]{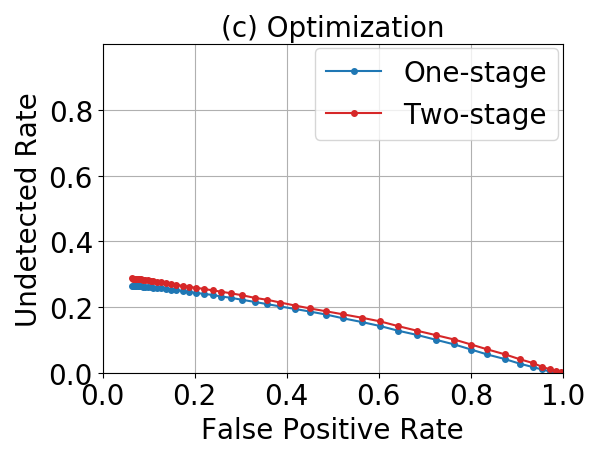}
\endminipage\hfill
\minipage{0.248\textwidth}
  \includegraphics[width=\linewidth]{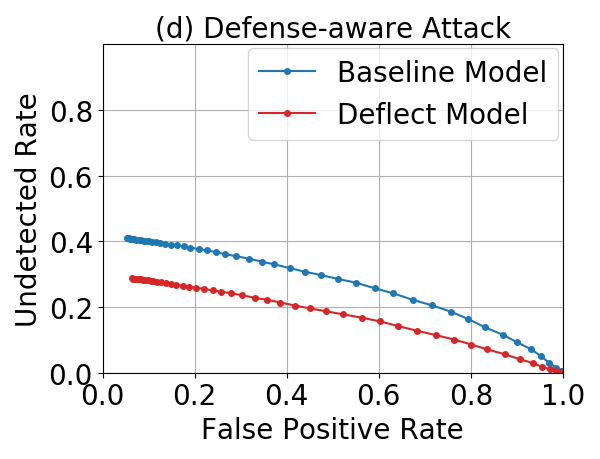}
\endminipage
\vspace{-3mm}
 \caption{(a) The success rate of white-box PGD and CC-PGD changes as the number of iterations increases for our deflecting model on CIFAR-10 dataset. (b) The success rate of white-box PGD and CC-PGD changes as $\epsilon_\infty$ increases for our deflecting model on CIFAR-10 dataset. (c) The Undetected Rate of the defense-aware attack CC-PGD optimized by a two-stage optimization and one-stage optimization vs. False Positive Rate for the clean data on the CIFAR-10 dataset. (d) Ablation study for cycle-consistency loss. The Undetected Rate of the defense-aware attack vs. False Positive Rate for baseline Capsule model trained without cycle-consistency loss and our deflecting model on the CIFAR-10 dataset. GTD and LBD are used to detect adversarial examples in baseline Capsule model. GTD, LBD and CCD are all used to detect adversarial attacks for our deflecting model.}\label{fig:sanity}
 \vspace{-4mm}
 \end{figure*}

\section{Experiments}
Now that we have proposed our new defense model, we first verify its detection performance on the SVHN and CIFAR-10 datasets on a variety of attacks. Then, we use a human study to demonstrate that our model frequently pressures the undetected attacks to be deflected.
\subsection{Evaluation Metrics and Datasets}
In this paper, we use \textbf{Accuracy} to represent the proportion of clean examples that are correctly classified by our network. We use \textbf{Success Rate} to measure the performance of an attack, which is defined as the proportion of adversarial examples that successfully fool the classifier into making the targeted prediction. In order to evaluate the performance of different detection mechanisms, we report both \textbf{False Positive Rate (FPR)}  and \textbf{Undetected Rate}. The False Positive Rate is the proportion of clean examples that are flagged as an adversarial example by the detection mechanism. The Undetected Rate, first proposed in~\cite{qin2019detecting}, denotes the proportion of adversarial examples that successfully fool the classifier and also go undetected. 
Finally, we perform a human study in Section~\ref{sec:hs} in order to show that our model is able to effectively deflect adversarial attacks. 

\subsection{Training Details and Test Accuracy}\label{sec:supp_train}
We set the batch size to be 64 and the learning rate to 0.0001 to train the network on SVHN. For CIFAR-10, the batch size is set to be 128 and the learning rate is 0.0002. We use the Adam optimizer~\cite{kingma2014adam} to train all models. The cycle-consistency loss $\ell
_{cyc}$ is empirically multiplied with 0.0005 before being added to the margin loss and the $\ell_2$ reconstruction loss used as in the original CapsNets~\cite{sabour2017}.

We test our deflecting models on the SVHN~\cite{netzer2011} and CIFAR-10 datasets~\cite{krizhevsky2009learning}. 
The classification accuracy on the clean test set is $96.5\%$ on SVHN and $92.6\%$ on CIFAR-10, which show that our deflecting models are reasonably good at classifying clean images.
\vspace{-3mm}
\subsection{Threat Model}
\vspace{-1mm}
In this paper, we consider two commonly used threat models: white-box and black-box. For white-box attacks, the adversary has full knowledge of the network architecture and parameters and is allowed to construct the adversarial attack by computing the gradient of model's output with respect to its input. In the black-box setting, the adversary is aware of the network architecture of the target model but does not have direct access to the model's parameters. To generate the black-box attacks against the target model, a substitute model that has the same network architecture is trained and further attacked by the white-box attacks, which are transferred to the target model as the black-box attacks.
\vspace{-3mm}
\subsection{Adversarial Attacks}
\vspace{-1mm}
Following the suggestions in~\cite{carlini2019evaluating}, we test our attack-agnostic detection mechanisms on three standard targeted attacks based on different distance metrics: $\ell_1$ norm-based EAD~\cite{chen2018ead}, $\ell_2$ norm-based CW~\cite{carlini2017towards}, and $\ell_\infty$ norm-based PGD~\cite{madry2017towards}.
In addition, we follow the suggestions in~\cite{carlini2017adversarial} to report the performance of our detection mechanisms against defense-aware attacks. We use CC-PGD (described in Section \ref{sec:ccpgd}) as our defense-aware attack. For the $\ell_\infty$ norm-based attacks, we set the maximal perturbation $\epsilon_\infty$ to be 16/255 on SVHN and 8/255 on CIFAR-10 as is typically used~\cite{buckman2018thermometer, madry2017towards}. 

To generate EAD and CW attacks, we follow the previous work~\cite{chen2018ead, carlini2017towards} to set the binary search steps to be 9, maximum iterations to be 1000 and learning rate to be 0.01. To construct $\ell_\infty$ norm-based attacks (PGD and our defense-aware CC-PGD), we use a step size 0.01 (2.55/255) in each iteration as~\cite{madry2017towards}. 
\vspace{-3mm}
\subsection{Sanity checks for PGD and CC-PGD attack}
\vspace{-1mm}
In this section, we perform basic sanity checks to ensure the adversarial attacks are correctly implemented and our proposed defense-aware CC-PGD is tuned well. In this section, we test attacks against our proposed deflecting model on the CIFAR-10 dataset. Similar conclusions also hold true on the SVHN dataset.
\vspace{-3mm}
\paragraph{Convergence of attacks.} Figure~\ref{fig:sanity} (a) shows the success rate of white-box PGD and CC-PGD varies as the number of iterations increases on the CIFAR-10 dataset.  We can see that the attacker has almost plateaued after 200 iterations. Therefore, we set the total number of attack steps to be 200 in generating PGD and CC-PGD attack for efficiency.
\vspace{-3mm}
\paragraph{100$\%$ success rate with non-constraint $\ell_\infty$ norm.}
In Figure~\ref{fig:sanity} (b), we show that the success rate of white-box PGD and CC-PGD varies as the $\ell_\infty$ bound of the adversarial perturbation $\epsilon_\infty$ increases. We can see that when $\epsilon_\infty$ is greater than 50/255, the success rate is 100$\%$. However, when $\epsilon_\infty$ is set to be 8/255 (which is typically used~\cite{buckman2018thermometer, madry2017towards}), the attack success rate against our deflecting model is below 50$\%$. 
\vspace{-3mm}
\paragraph{Two-stage optimization}
To demonstrate the effectiveness of our used two-stage optimization in generating defense-aware CC-PGD, we compare the attack performance of two-stage optimization introduced in Section~\ref{sec:ccpgd} and a one-stage optimization that uses a single loss function which combines the cross-entropy loss to fool the classifier with the reconstruction loss $\ell_r$ in Eqn.~\ref{loss} to fool the detectors. In Figure~\ref{fig:sanity} (c), we construct the defense-aware CC-PGD against our deflecting model on the CIFAR-10 dataset using one-stage and two-stage optimization respectively. We can see that the defense-aware CC-PGD attack that is optimized by the two-stage optimization is slightly better than that optimized by the one-stage optimization. Therefore, we follow~\cite{qin2019detecting} to use the two-stage optimization in all the following experiments to construct CC-PGD attack.
\vspace{-7mm}
\paragraph{Hyperparameters}
Empirically, the hyperparameter $\alpha_1$, $\alpha_2$ and $\alpha_3$ in Eqn.~\ref{loss} are set to be 1, 0 and 20 respectively to balance the importance among three detectors in generating our defense-aware CC-PGD. Since the Cycle-Consistency Detector is the most effective detector (discussed below in Section~\ref{sec:detect}), we assign a much higher weight to $\alpha_3$, which controls the importance of attacking Cycle-Consistency Detection in generating our defense-aware CC-PGD attack. In addition, we observe that increasing $\alpha_2$ (controlling the importance of attacking the Local Best Detector) leads to a decrease of the attack performance). Therefore, $\alpha_2$ is set to be 0. This might result from the contradiction between minimizing the winning-capsule reconstruction and maximizing the losing-capsule reconstruction, where they share the background capsule information. Lastly, $\alpha_1$ is set to be a very small value as 1 for the best attack performance for CC-PGD. 

The parameter that balances the importance of the two stages in CC-PGD is empirically set to be 0.5 on SVHN and 0.75 for the first stage and 0.25 for the second stage on CIFAR-10. More detailed results about selecting these hyperparameters are shown in Supplementary Material.

\begin{figure}[t]
\minipage{0.24\textwidth}
  \includegraphics[width=\linewidth]{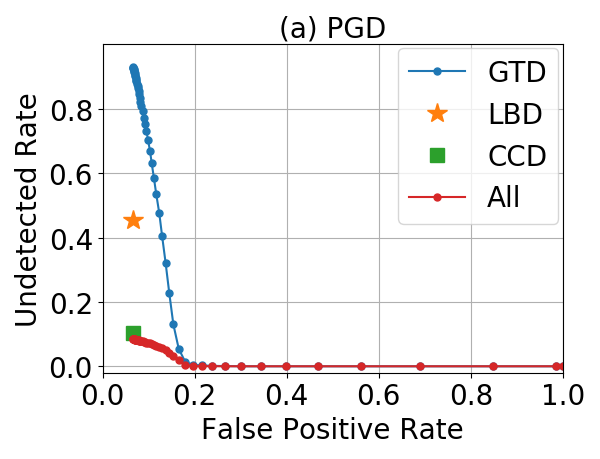}
\endminipage
\minipage{0.24\textwidth}%
  \includegraphics[width=\linewidth]{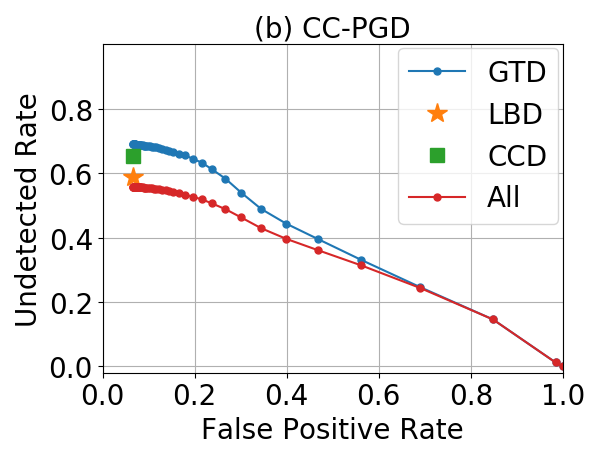}
\endminipage
\vspace{-5mm}
 \caption{The Undetected Rate of different detectors for white-box attacks versus False Positive Rate (FPR) for clean input on the SVHN dataset. ``All'' denotes GTD, LBD and CCD are all used to detect adversarial attacks. The testing model is our deflecting model. The better detection mechanism has a smaller FPR for clean input and smaller undetected rate for attacks.}\label{fig:ablation}
 \vspace{-4mm}
 \end{figure}

\vspace{-2mm}
\subsection{Ablation Study}
\vspace{-1mm}
\subsubsection{Detection methods}\label{sec:detect}
In this section, we study the effectiveness of our proposed detection mechanisms: Local Best Detector (LBD) and Cycle-Consistency Detector (CCD) and compare them with Global Threshold Detector (GTD) from ~\cite{qin2019detecting}. 

Since the False Positive Rate (FPR) of clean input flagged by the Global Threshold Detector (GTD) varies as the chosen global threshold, in Figure~\ref{fig:ablation} we plot the undetected rate of white-box adversarial attacks flagged by different detectors versus the False Positive Rate (FPR) of the clean input. The global threshold $\theta$ is chosen from the range [0, 20] with a step size of 0.4. 
We can clearly see that: 1) A single Global Threshold Detector (GTD) proposed in~\cite{qin2019detecting} is not enough to effectively detect adversarial attacks. 2) In a standard PGD attack, the CCD is the most effective detector at a low False Positive Rate, similar conclusions on EAD and CW attacks shown in Supplementary Material. 
3) In all the attacks, the combination of all three detectors always performs the best. Therefore, we only report the performance of the undetected rate of the combination of all three detectors in the following experiments.
\begin{figure*}[t]
\centering
\minipage{0.5\textwidth}
\centering
  \includegraphics[width=0.3\linewidth]{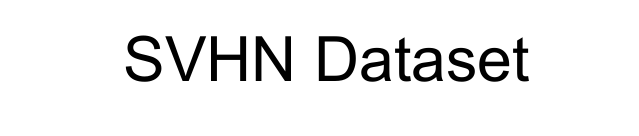}

\endminipage
\minipage{0.5\textwidth}
\centering
  \includegraphics[width=0.38\linewidth]{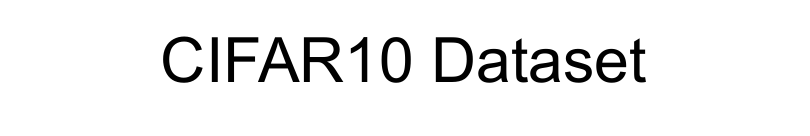}
\endminipage\\
\minipage{0.25\textwidth}
  \includegraphics[width=\linewidth]{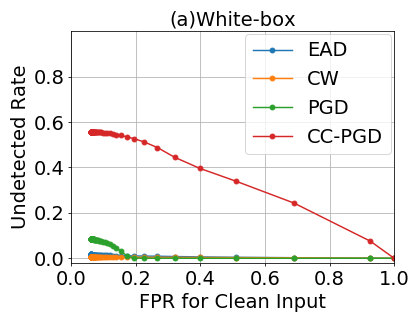}
\endminipage\hfill
\minipage{0.25\textwidth}
  \includegraphics[width=\linewidth]{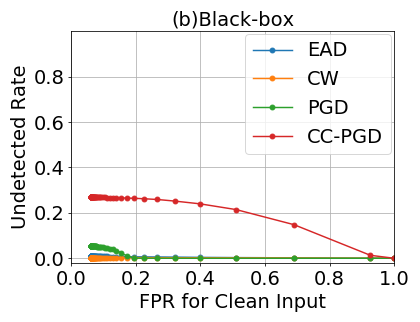}
\endminipage\hfill
\minipage{0.25\textwidth}
  \includegraphics[width=\linewidth]{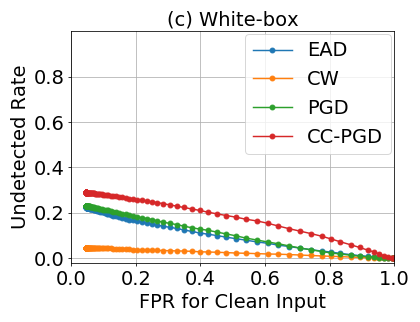}
\endminipage\hfill
\minipage{0.25\textwidth}%
  \includegraphics[width=\linewidth]{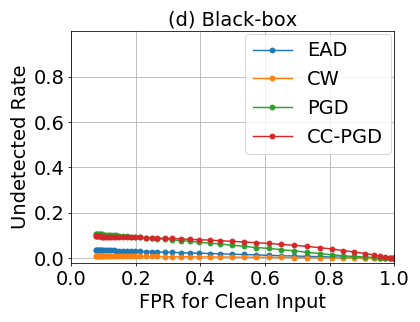}
\endminipage
\vspace{-5mm}
 \caption{The Undetected Rate for white-box and black-box attacks versus False Positive Rate (FPR) for clean input on the SVHN and CIFAR-10 datasets. The strongest attack has the largest area under the line.
}\label{fig:perf}
\vspace{-2mm}
\end{figure*}

\begin{table*}[t]
\vspace{-1mm}
\centering
\caption{Success rate of the white-box and black-box attacks for our deflecting model.}\label{tab:success_rate}
\begin{tabu} to 0.98\textwidth {c|X[c]X[c]|X[c]X[c]|X[c]X[c]|X[c]X[c]}
\toprule
\multirow{2}{*}{Dataset} &  \multicolumn{2}{c|}{EAD} & \multicolumn{2}{c|}{CW} & \multicolumn{2}{c|}{PGD} & \multicolumn{2}{c}{CC-PGD} \\ 
                                                 & White       & Black      & White      & Black      & White       & Black      & White        & Black        \\ \midrule
SVHN    
              & 100.0\%             &     10.1\%                  & 97.6\%       & 1.7\%        & 96.0\%        & 28.7\%       & 69.0\%         & 37.0\%         \\\midrule
CIFAR-10 
&    100.0\%          &      6.9\%     &      78.0\%      &    1.6\%         &   49.3\%           &   15.5\%          &     46.8\%          &     12.9\%          \\ \bottomrule
\end{tabu}
\vspace{-4mm}
\end{table*}
\vspace{-3mm}
\subsubsection{Cycle-consistency loss}
\vspace{-1mm}
To demonstrate the effectiveness of the proposed cycle-consistency loss, we construct a baseline Capsule model that has the same network architecture as our deflecting model but is trained without the extra cycle-consistency loss. The False Positive Rate of the Cycle-Consistency Detector on the CIFAR-10 test set is 33.46$\%$, which represents that 33.46$\%$ of the clean test images are incorrectly flagged as an adversarial example by the Cycle-Consistency Detector. This means the Cycle-Consistency Detector is not suitable for a model that is trained without cycle-consistency loss. Therefore, to compare the detection performance between the baseline Capsule model and our deflecting model, we use a combined Global Threshold Detector (GTD) and Local Best Detector (LBD) for the baseline Capsule model and all three detectors for the deflecting model. The undetected rate of the white-box defense-aware attack versus the False Positive Rate (FPR) of the clean input on the CIFAR-10 dataset is shown in Figure~\ref{fig:sanity} (d), where we can see that our deflecting model together with all three detectors has a better detection performance compared to the baseline model trained without the cycle-consistency loss. 
\vspace{-3mm}
\subsection{Detection of White-box Attacks}
\vspace{-1mm}
Before showing that our defense produces deflected attacks, we must first validate that it improves detection performance. Therefore, we test our model on standard and defense-aware attacks and compare it with state-of-the-art detection methods in this section.
\vspace{-3mm}
\paragraph{Standard attacks} As shown in Figure~\ref{fig:perf}, our detection method has a very small undetected rate for all three standard white-box attacks (EAD, CW and PGD) on both the SVHN and CIFAR-10 dataset. Among them, PGD is the strongest attack against our detection mechanisms with the highest undetected rates at the same FPR. For PGD attacks, we achieve an undetected rate below 10$\%$ with a small False Positive Rate on the SVHN dataset. The undetected rate for white-box PGD is around $22\%$ with the smallest False Positive Rate on the CIFAR-10 dataset. These demonstrate that our detection mechanism is very effective in detecting standard white-box attacks that are based on different $\ell_p$ norms. 
\vspace{-3mm}
\paragraph{Defense-aware attacks} Following the suggestions in~\cite{carlini2017adversarial}, we test our detection mechanism in the setting where the adversary is fully aware of the defense (``defense-aware attacks'') using the CC-PGD attack. Since the PGD attack is stronger than EAD and CW in attacking our deflecting model (shown in Figure~\ref{fig:perf}), the first stage of our CC-PGD attack is to construct an adversarial image via standard PGD and then, in the second stage, take the reconstruction error and cycle-consistency into consideration in order to fool the detection methods. In Figure~\ref{fig:perf} we can clearly see the undetected rate of CC-PGD increases compared to a standard PGD attack. However, there is a significant performance drop in the success rate of White-box CC-PGD (from PGD: 96.0$\%$ to CC-PGD: 69.0$\%$ on  SVHN) as shown in Table~\ref{tab:success_rate}. This indicates that the adversary needs to sacrifice some success rate in order not to be detected by our detection mechanism.
\vspace{-3mm}
\paragraph{Comparison with State-of-the-Art Detection Methods}
\begin{table}[t]
\vspace{-2.2mm}
\caption{Comparison of the Undetected Rate of the state-of-the-art detection methods on the CIFAR-10 dataset. For all the models, the maximum $\ell_\infty$ perturbation is $\epsilon_\infty = 8/255$ of the pixel dynamic range and the False Positive Rate of the clean input are 5$\%$. The best detection performance are highlighted in \textbf{bold}. (Smaller numbers indicate better detection performance.)}\label{tab:compare}
\begin{tabu}{c|c|c|c}
\toprule
{Detection Method} & {Statistical} & {Classifier-}     & Ours  \\
 & Test & based & \\ \midrule
CW & 0.1\% & \textbf{0.0}\%   & 4.6\% \\

Defense-aware PGD& 97.8\%            & 98.4\%                         &    \textbf{28.9}\%                           \\ \bottomrule
\end{tabu}
\vspace{-3mm}
\end{table}
We compare our detection methods with the most recent statistical test-based detection method~\cite{roth2019odds} and a classifier-based detection method proposed  in~\cite{hosseini2019odds}. 
In Table~\ref{tab:compare}, we can see that although the statistical test~\cite{roth2019odds} and the classifier-based detection method~\cite{hosseini2019odds} can detect standard attacks successfully, they both fully fail against defense-aware attacks \vspace{1mm}\footnote{The numbers of statistical test and classifier-base detection in the Table~\ref{tab:compare} are extracted from~\cite{hosseini2019odds}. Since the success rate of the attacks are close to 100$\%$, the undetected rate is roughly (1 - True Positive Rate).}. In contrast, our proposed reconstruction-based detection mechanism has the best undetected rate in detecting defense-aware adversarial attacks and a very small undetected rate of $4.6\%$ in detecting CW attacks. 
\begin{figure*}[t]
\begin{minipage}{0.33\linewidth}
\centering
\includegraphics[width=\linewidth]{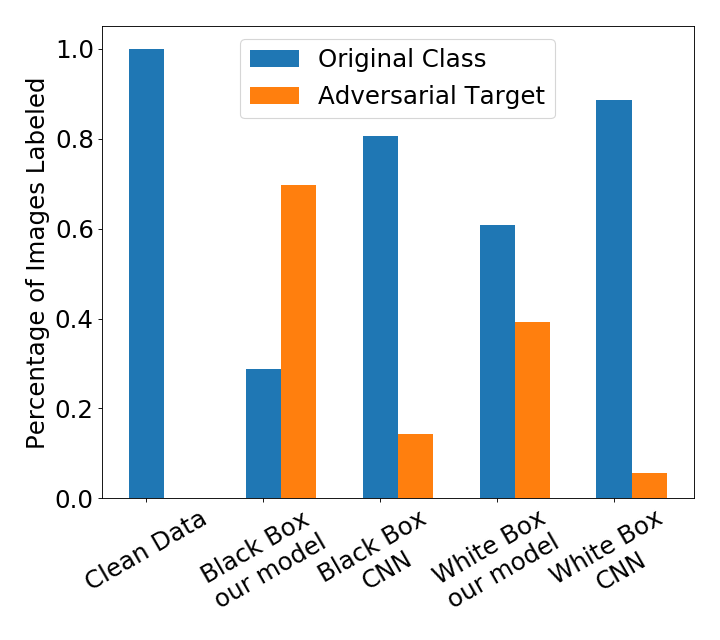}
\vspace{-9mm}
\caption{The human study results on SVHN. The maximal $\ell_\infty$ perturbation is 16/255.}
\label{fig:human}
\end{minipage}\hfill
\begin{minipage}{0.65\linewidth}
\centering
\includegraphics[width=0.94\linewidth]{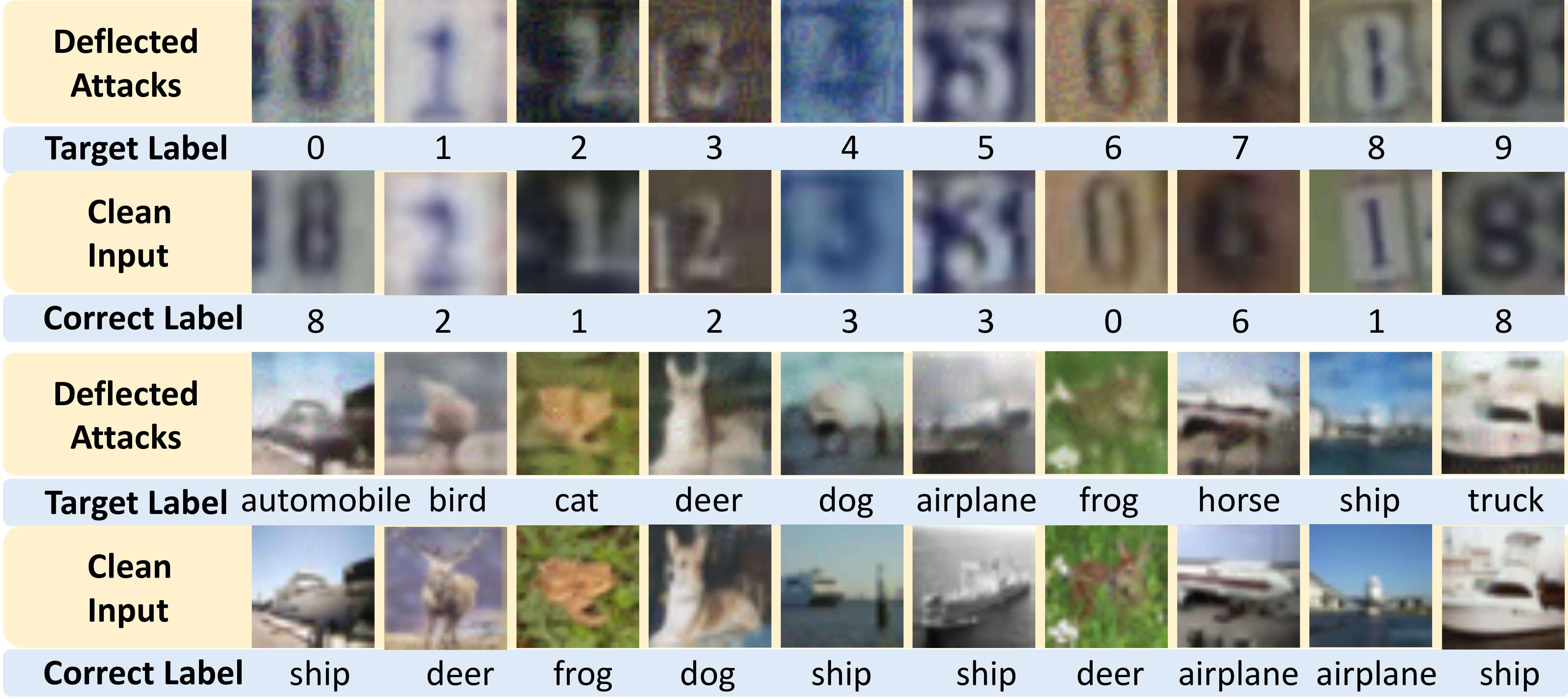}
\vspace{-2mm}
\caption{Deflected adversarial attacks on SVHN and CIFAR-10. The maximal $\ell_\infty$ perturbation is 16/255 for SVHN and 25/255 for CIFAR-10.}
\label{fig:cifar}
\end{minipage}
\vspace{-5mm}
\end{figure*}
  \vspace{-1mm}
\subsection{Detection of black-box Attacks}
To study the effectiveness of our detection mechanisms, we also test our models on black-box attacks. In Figure~\ref{fig:perf} we can see that  the undetected rate when the inputs are black-box CC-PGD attacks is only half of that for white-box CC-PGD on both datasets. The highest undetected rate of a black-box attack is around $13\%$ on the CIFAR-10 dataset, which demonstrates that our detection mechanism can successfully detect black-box defense-aware attacks. In addition, the great gap of the success rate between white-box and black-box attacks shown in Table~\ref{tab:success_rate} indicates our defense model significantly reduces the transferability of all kinds of adversarial attacks.
\vspace{-2mm}
\section{Deflected Attacks}\label{sec:hs}
The numbers that we presented earlier in this paper have implicitly assumed all adversarial attacks still resemble the initial class, and therefore classifying them as the target class would constitute a mistake. This assumption may not be true in practice. We have discussed the ability of our model to deflect adversarial attacks by having adversarial gradients aligned with the class conditional data distribution, thereby making adversarial attacks resemble the target class. To quantify these claims we need to evaluate human performance on the adversarial attacks against our model. 
\subsection{Human Study on SVHN}
\vspace{-1mm}
To validate our claim that our method can deflect adversarial attacks, we performed a human study by using the Amazon Mechanical Turk web service to recruit participants and asked people to label SVHN digits. Each time, they were shown a single image which was randomly sampled from the following five different sets: 1) clean images from the SVHN test set, 2) and 3) the undetected and successful black-box and white-box PGD and CC-PGD adversarial attacks against our deflecting model, 4) and 5) the successful black-box and whilte-box PGD attacks generated to attack a standard CNN classifier\footnote{The CNN classifier has the same network architecture as the classification network in our deflecting model except that we replace the CapsLayer with a convolutional layer.}. The maximal adversarial perturbation of all the $\ell_\infty$ norm-based attacks are bounded by the same $\epsilon_\infty=16/255$. The recruiters were asked to classify each image as a digit between 0 and 9. If multiple digits occurred in one image, we asked people to label the digit closest to the center of the image. We did not limit the labelling time did not explain the purpose of this study to the users other than it was a research study. In this way, we had 1500 images labeled in total and each image was labeled by five different users. We then calculated the percentage of uniformly labeled images that were classified as either the original class or the adversarial target class. The results are summarized in Figure~\ref{fig:human}. 

We can see that 69.7\% of successful and undetected black-box attacks against our model were classified as the adversarial target. This means that when our defense is attacked with adversarial attacks generated within a standard $\ell_\infty$ bound, not only are the results visibly different than the source image, they resemble the target class. In this way, these attacks are successfully deflected and can hardly be said to be adversarial, as the network is classifying them the same way our human testers classified them. This is not the case for the baseline CNN model, where only 14.3\% of the successful black-box PGD attacks were labeled as the target class. In addition, compared to the white-box attacks, more undetected and successful adversarial attacks generated under the black-box setting are deflected to resemble the target class. This suggests that to attack our deflecting model in a more practical setting (black-box), the attack ends up being deflected in order not to be detected, as shown in Figure~\ref{fig:cifar}.
\vspace{-3mm}
\subsection{Deflected Attacks on CIFAR-10}
\vspace{-1mm}
To show that our model can effectively deflect adversarial attacks on the CIFAR-10 dataset, we have chosen a deflected adversarial attack for each class with a maximal $\ell_\infty$ norm as 25/255, displayed in Figure~\ref{fig:cifar}. It is apparent that the clean input has been perturbed to have the representative features of the target class, in order to fool both the classifier and our detection mechanisms. 
As a result, these adversarial attacks are also successfully deflected by our model. Unlike SVHN, for which human evaluators reliably classified the attacks as the target label, the generated adversarial attacks against our deflecting model on the CIFAR-10 do not reliably resemble the target class, though they are much harder to identify than the clean data.
\vspace{-3mm}
\section{Conclusion}
\vspace{-1mm}
In this paper, we introduce a new approach that presents a step towards ending the battle between defenses and attacks by deflecting adversarial attacks. To this end, we propose a new cycle-consistency loss to encourage the winning-capsule reconstruction of the CapsNet to closely match the class-conditional distribution. 
With three detection mechanisms, we are able to detect standard adversarial attacks based on three different distance metrics with a low False Positive Rate on SVHN and CIFAR-10. To specifically attack our detection mechanisms, we propose a defense-aware attack and find that our model achieves drastically lower undetected rates for defense aware attacks compared to state-of-the-art methods. In addition, a large percentage of the undetected attacks are deflected by our model to resemble the adversarial target class, stop being adversarial any more. 
This is verified by a human study showing that 70\% of the undetected black-box adversarial attacks are classified unanimously by humans as the target class on SVHN.

\bibliographystyle{icml2020}

\newpage
\section*{Appendix}
\appendix
\begin{table*}[h!]
\centering
\caption{The network architecture for the SVHN dataset.}\label{tab:arch_svhn}
\begin{tabular}{|c|c|l|}
\hline
& Layer Name      & \multicolumn{1}{c|}{Configurations}                                                                                                                                        \\ \hline
\multirow{9}{*}{\begin{tabular}[c]{@{}c@{}}Classification \\ Network\end{tabular}} & Conv            & \begin{tabular}[c]{@{}l@{}}filter size: 3x3, number of filters: 64x4, stride size: 1x1, \\ activation: leaky relu\end{tabular}  
                                           \\ \cline{2-3} 
                                           & Conv            
         & \begin{tabular}[c]{@{}l@{}}filter size: 3x3, number of filters: 64x8, stride size: 1x1, \\ activations: leaky relu\end{tabular}                                            \\ \cline{2-3} 
& Avg Pooling     & pool size: 2x2, stride size: 2x2                                                                                                                                           \\ \cline{2-3} 
&Conv            & \begin{tabular}[c]{@{}l@{}}filter size: 3x3, number of filters: 64x2, stride size: 1x1, \\ activation: leaky relu\end{tabular}                                             \\ \cline{2-3} 
       & Conv            & \begin{tabular}[c]{@{}l@{}}filter size: 3x3, number of filters: 64x4, stride size: 1x1, \\ activation: leaky relu\end{tabular}                                             \\ \cline{2-3} 
& Avg Pooling     & pool size: 2x2, stride size: 2x2                                                                                                                                           \\ \cline{2-3} 
& Conv            & \begin{tabular}[c]{@{}l@{}}filter size: 3x3, number of filters: 64x1, stride size: 1x1, \\ activation: leaky relu\end{tabular}                                               \\ \cline{2-3}           & Conv            & \begin{tabular}[c]{@{}l@{}}filter size: 3x3, number of filters: 64x2, stride size: 1x1, \\ activation: leaky relu\end{tabular}                                             \\ \cline{2-3}              & CapsLayer       & \begin{tabular}[c]{@{}l@{}}number of input capsules: 16, input atoms: 512, \\ number of output capsules: 25, output atoms: 4, \\ number of dynamic routing: 1\end{tabular} \\ \hline
\multirow{5}{*}{\begin{tabular}[c]{@{}c@{}}Reconstruction\\ Network\end{tabular}}  & fully connected & input size: 100, output size:1024                                                                                                                                          \\ \cline{2-3}             & fully connected & input size: 1024, output size:16384                             \\ \cline{2-3}               & deconv          & filter size: 4x4, number of filters: 64, stride size: 2x2                                                                                                                  \\ \cline{2-3}              & deconv          & filter size: 4x4, number of filters: 32, stride size: 2x2                                                                                                                  \\ \cline{2-3}              & conv            & \begin{tabular}[c]{@{}l@{}}filter size: 4x4 number of filters: 3, stride size: 1x1, \\ activation: sigmoid
\end{tabular}   \\ \hline
\end{tabular}
\end{table*}

\begin{table*}[h]
\centering
\caption{The network architecture for the CIFAR-10 dataset.}\label{tab:arch_cifar}
\begin{tabular}{|c|c|l|}
\hline
                                                                                   & Layer Name      & \multicolumn{1}{c|}{Configurations}                                                              {}                                                                          \\ \hline
\multirow{9}{*}{\begin{tabular}[c]{@{}c@{}}Classification \\ Network\end{tabular}} & Conv            & \begin{tabular}[c]{@{}l@{}}filter size: 3x3, number of filters: 128x4, stride size: 1x1, \\ activation: leaky relu\end{tabular}                                             \\ \cline{2-3} 
                                                                                   & Conv            & \begin{tabular}[c]{@{}l@{}}filter size: 3x3, number of filters: 128x8, stride size: 1x1, \\ activations: leaky relu\end{tabular}                                            \\ \cline{2-3} 
                                                                                   & Avg Pooling     & pool size: 2x2, stride size: 2x2                                                                                                                                           \\ \cline{2-3} 
                                                                                   & Conv            & \begin{tabular}[c]{@{}l@{}}filter size: 3x3, number of filters: 128x2, stride size: 1x1, \\ activation: leaky relu\end{tabular}                                             \\ \cline{2-3} 
                                                                                   & Conv            & \begin{tabular}[c]{@{}l@{}}filter size: 3x3, number of filters: 128x4, stride size: 1x1, \\ activation: leaky relu\end{tabular}                                             \\ \cline{2-3} 
                                                                                   & Avg Pooling     & pool size: 2x2, stride size: 2x2                                                                                                                                           \\ \cline{2-3} 
                                                                                   & Conv            & \begin{tabular}[c]{@{}l@{}}filter size: 3x3, number of filters: 128x1, stride size: 1x1, \\ activation: leaky relu\end{tabular}                                               \\ \cline{2-3} 
                                                                                   & Conv            & \begin{tabular}[c]{@{}l@{}}filter size: 3x3, number of filters: 128x2, stride size: 1x1, \\ activation: leaky relu\end{tabular}                                             \\ \cline{2-3} 
                                                                                   & CapsLayer       & \begin{tabular}[c]{@{}l@{}}number of input capsules: 16, input atoms: 512, \\ number of output capsules: 25, output atoms: 8, \\ number of dynamic routing: 1\end{tabular} \\ \hline
\multirow{5}{*}{\begin{tabular}[c]{@{}c@{}}Reconstruction\\ Network\end{tabular}}  & fully connected & input size: 200, output size:1024                                                                                                                                          \\ \cline{2-3} 
                                                                                   & fully connected & input size: 1024, output size:16384                                                                                                                                        \\ \cline{2-3} 
                                                                                   & deconv          & filter size: 4x4, number of filters: 64, stride size: 2x2                                                                                                                  \\ \cline{2-3} 
                                                                                   & deconv          & filter size: 4x4, number of filters: 32, stride size: 2x2                                                                                                                  \\ \cline{2-3} 
                                                                                   & conv            & \begin{tabular}[c]{@{}l@{}}filter size: 4x4 number of filters: 3, stride size: 1x1, \\ activation: sigmoid\end{tabular}                                                    \\ \hline
\end{tabular}
\end{table*}

\section{Model Architectures}\label{sec:supp_arch}
The details of the network architecture for SVHN~\cite{netzer2011} and CIFAR-10~\cite{krizhevsky2009learning} datasets are shown in Table~\ref{tab:arch_svhn} and Table~\ref{tab:arch_cifar}.

\section{Ablation Study for Detection Methods}
In Figure~\ref{fig:ablation}, we show the undetected rate of white-box EAD~\cite{chen2018ead} and CW~\cite{carlini2017towards} attacks flagged by different detectors versus the False Positive Rate (FPR) of the clean input. The combination of three detectors always works the best in detecting adversarial examples.
\begin{figure}[h]
\minipage{0.24\textwidth}
  \includegraphics[width=\linewidth]{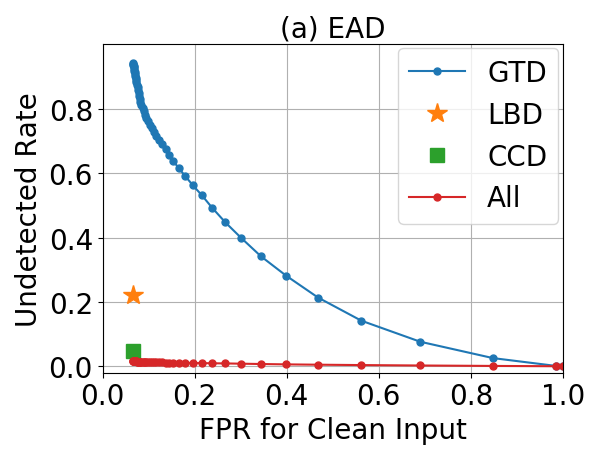}
\endminipage
\minipage{0.24\textwidth}
  \includegraphics[width=\linewidth]{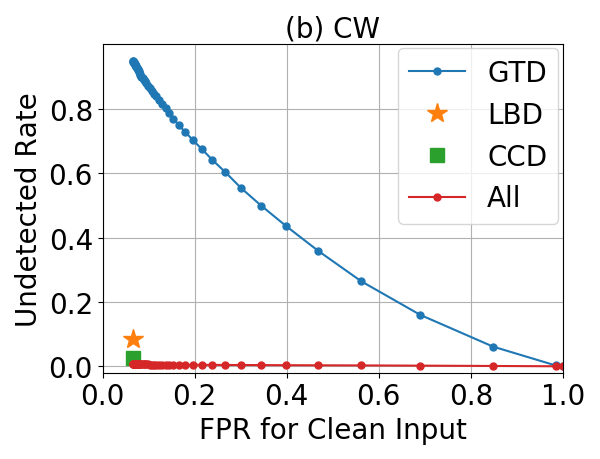}
\endminipage\hfill
 \caption{The Undetected Rate of different detectors for white-box attacks versus False Positive Rate (FPR) for clean input on the SVHN dataset. ``All'' denotes GTD, LBD and CCD are all used to detect adversarial attacks. The testing model is our deflecting model. The better detection mechanism has a smaller FPR for clean input and smaller undetected rate for attacks.}\label{fig:ablation}
 \end{figure}

 \section{Hyperparameters}
\begin{figure}
 \centering
 \includegraphics[width=\linewidth]{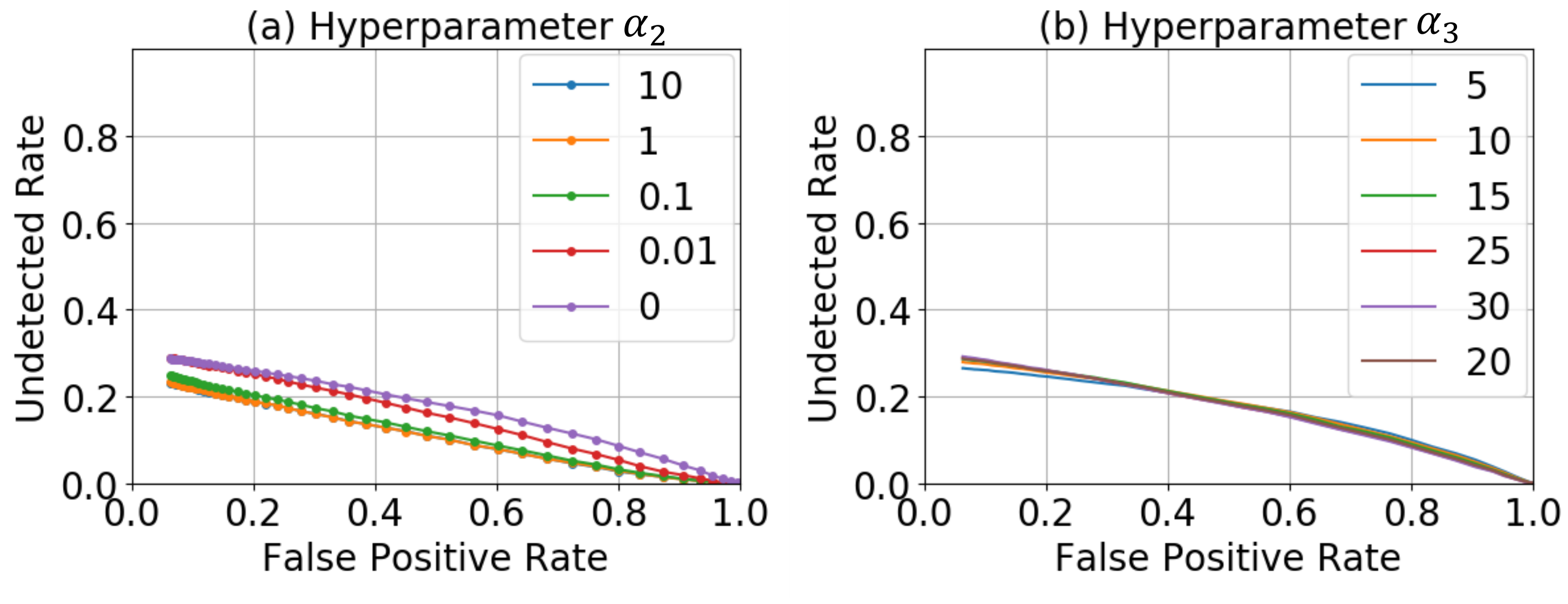}
 \vspace{-5mm}
 \caption{The undetected rate of our white-box defense-aware CC-PGD attack versus False Positive Rate (FPR) for clean input on the CIFAR-10 dataset when we change the hyperparameter $\alpha_2$ in (a) and hyperparameter $\alpha_3$ in (b). These hyperparameters control the importance of attacking each detector in Eqn.~\ref{loss}.}\label{param}
 \end{figure}

As introduced in Section 4 in the main paper, we construct the defense-aware CC-PGD attack via minimizing the reconstruction loss, which is defined as:
\begin{equation}\label{loss}
\begin{split}
\ell_{r}(\vx') & = \alpha_1 \cdot  \ell_{g}(\vx') + \alpha_2 \cdot \ell_{l}(\vx') + \alpha_3 \cdot \ell_{cyc}(\vx')\\
& = \alpha_1 \cdot \lVert r(\vv_{i=f(\vx')}) - \vx' \rVert_2  \\
& \quad \quad - \alpha_2 \cdot  \frac{\sum_{k \neq f(\vx')}^{n}\lVert r(\vv_{k}) - \vx' \rVert_2}{n-1} \\ 
&  \quad \quad \quad \quad + \alpha_3 \cdot \ell_{net}(f(r(\vv_{i=f(\vx')})), f(\vx'))
\end{split}
\end{equation}
where $\vx' = \vx+\Delta$ is the adversarial example, $n$ is the number of the classes in the dataset, $\lVert r(\vv_{i=f(\vx')}) - \vx' \rVert_2$ is the winning-capsule reconstruction error and $\lVert r(\vv_{k\neq f(\vx')}) - \vx' \rVert_2$ is the losing-capsule reconstruction error. The hyperparameters $\alpha_1$, $\alpha_2$ and $\alpha_3$ are used to balance the importance of attacking each detector. We set $\alpha_1=1$ and then show the attack performance when we change $\alpha_2$ (see Figure~\ref{param} (a)) and $\alpha_3$ (see Figure~\ref{param} (b)).

We can see that when we set $\alpha_2 = 0$, the attack performance is the best (higher undetected rate at a low False Positive Rate). In addition, the attack performance of our CC-PGD is not sensitive to the hyperparameter $\alpha_3$. Therefore, we simply set $\alpha_3=20$, which is slightly better at a low False Positive Rate.
\section{Examples of Adversarial Attacks and Reconstructions}
\begin{figure*}[]
\begin{minipage}{0.49\linewidth}
  \includegraphics[width=1\linewidth]{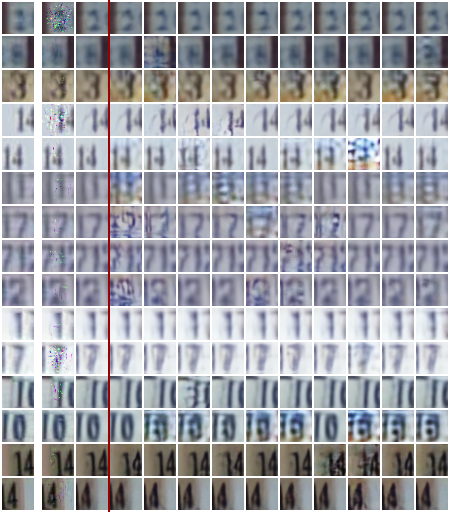}
  \end{minipage}\hfill
   \begin{minipage}{0.49\linewidth}
  \includegraphics[width=1\linewidth]{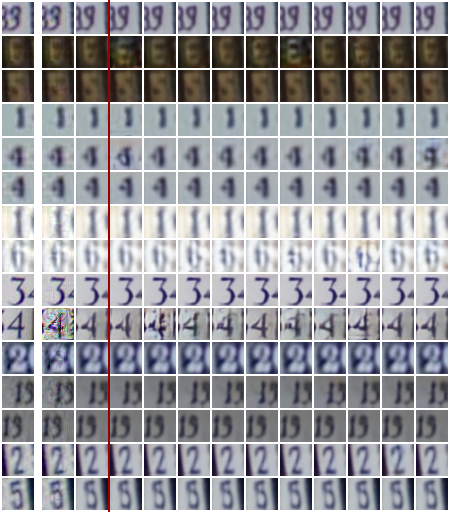}
\end{minipage}
\caption{Successful but detected adversarial EAD attacks (on the left) and CW attacks (on the right) and the corresponding capsule reconstructions on SVHN. The first column is the clean input, the second column is the adversarial example, the third column is the winning-capsule reconstruction, the last ten columns are the reconstructions corresponding to class 0 to 9.}\label{fig:svhn_cw}
\end{figure*}

\begin{figure*}[]
\begin{minipage}{0.49\linewidth}
  \includegraphics[width=1\linewidth]{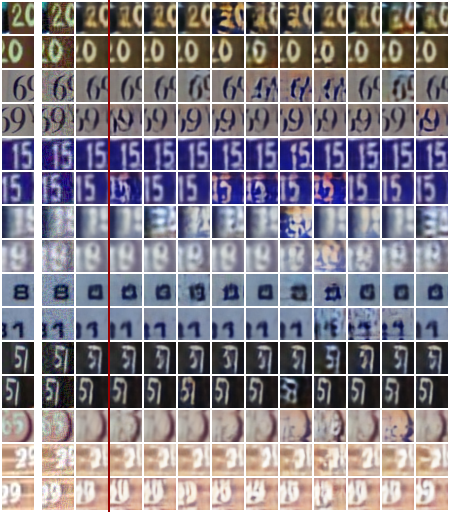}
  \end{minipage}\hfill
   \begin{minipage}{0.49\linewidth}
  \includegraphics[width=1\linewidth]{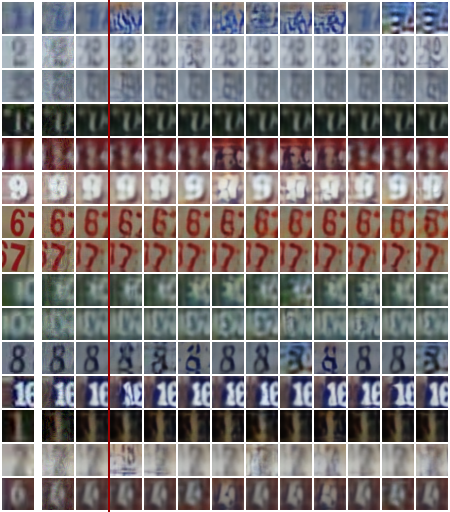}
\end{minipage}
\caption{Successful but detected adversarial PGD attacks (on the left) and our CC-PGD attacks (on the right) and the corresponding capsule reconstructions on SVHN. The first column is the clean input, the second column is the adversarial example, the third column is the winning-capsule reconstruction, the last ten columns are the reconstructions corresponding to class 0 to 9. The maximal $\ell_\infty$ bound to the adversarial perturbation is 16/255.}\label{fig:svhn_pgd}
\end{figure*}

\begin{figure*}[]
\begin{minipage}{0.49\linewidth}
  \includegraphics[width=1\linewidth]{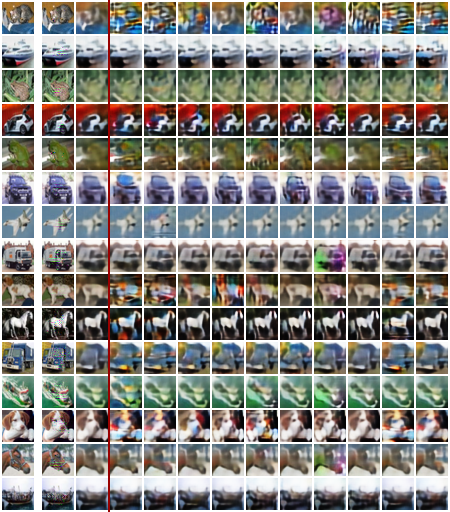}
  \end{minipage}\hfill
   \begin{minipage}{0.49\linewidth}
  \includegraphics[width=1\linewidth]{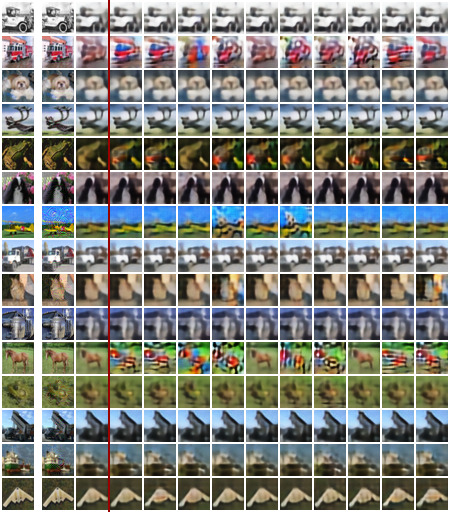}
\end{minipage}
\caption{Successful but detected adversarial EAD attacks (on the left) and CW attacks (on the right) and the corresponding capsule reconstructions on CIFAR-10. The first column is the clean input, the second column is the adversarial example, the third column is the winning-capsule reconstruction, the last ten columns are the reconstructions corresponding to class 0 to 9.}\label{fig:cifar_cw}
\end{figure*}

\begin{figure*}[]
\begin{minipage}{0.49\linewidth}
  \includegraphics[width=1\linewidth]{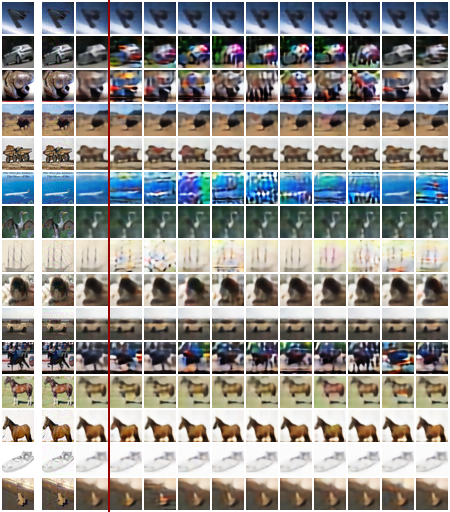}
  \end{minipage}\hfill
   \begin{minipage}{0.49\linewidth}
  \includegraphics[width=1\linewidth]{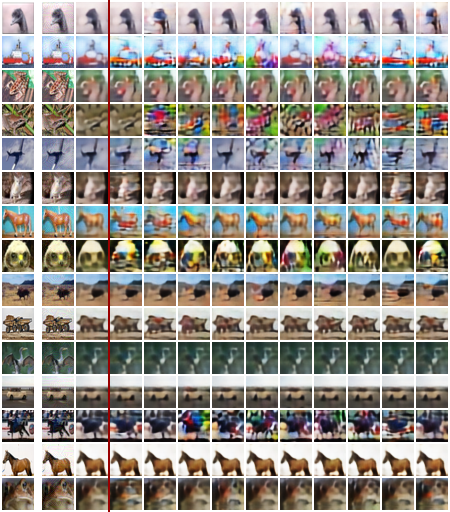}
\end{minipage}
\caption{Successful but detected adversarial PGD attacks (on the left) and our CC-PGD attacks (on the right) and the corresponding capsule reconstructions on CIFAR-10. The first column is the clean input, the second column is the adversarial example, the third column is the winning-capsule reconstruction, the last ten columns are the reconstructions corresponding to class 0 to 9. The maximal $\ell_\infty$ bound to the adversarial perturbation is 8/255.}\label{fig:cifar_pgd}
\end{figure*}

We display successful adversarial attacks but detected by our detection mechanism, and display all the reconstructions when the input are EAD attacks (on the left) and CW attacks (on the right) in Figure~\ref{fig:svhn_cw}, PGD attacks (on the left) and our CC-PGD attacks (on the right) in Figure~\ref{fig:svhn_pgd} for the SVHN dataset. We also show the successful and detected adversarial EAD attacks (on the left) and CW attacks (on the right) in Figure~\ref{fig:cifar_cw}, PGD attacks (on the left) and our CC-PGD attacks (on the right) in Figure~\ref{fig:cifar_pgd} for CIFAR-10 dataset.
\newpage

\end{document}